\documentclass[conference]{IEEEtran}

\usepackage{cite}
\usepackage{graphicx}
\usepackage[cmex10]{amsmath}
\usepackage{amsfonts}
\usepackage{url}
\usepackage{multirow}

\begin{document}
%
\title{Two-Stream Thermal Imaging Fusion for\\ Enhanced Time of Birth Detection in Neonatal Care }

\author{\IEEEauthorblockN{Jorge García-Torres\IEEEauthorrefmark{1},
Øyvind Meinich-Bache\IEEEauthorrefmark{1}\IEEEauthorrefmark{2},
Sara Brunner\IEEEauthorrefmark{2}, 
Siren Rettedal\IEEEauthorrefmark{3}\IEEEauthorrefmark{4},
Vilde Kolstad\IEEEauthorrefmark{4} and
Kjersti Engan\IEEEauthorrefmark{1}\IEEEauthorrefmark{5}}
\\
\IEEEauthorblockA{\IEEEauthorrefmark{1}Department of Electrical Engineering and Computer Science, University of Stavanger, Norway}
\IEEEauthorblockA{\IEEEauthorrefmark{2}Strategic Research, Laerdal Medical, Stavanger, Norway}
\IEEEauthorblockA{\IEEEauthorrefmark{3}Faculty of Health Sciences, University of Stavanger, Norway}
\IEEEauthorblockA{\IEEEauthorrefmark{4}Department for Simulation-based Learning, Stavanger University Hospital, Norway}
\IEEEauthorblockA{\IEEEauthorrefmark{5}Corresponding author: kjersti.engan@uis.no}\\} 

\maketitle

\begin{abstract}
Around 10\% of newborns require some help to initiate breathing, and 5\% need ventilation assistance.
Accurate Time of Birth (ToB) documentation is essential for optimizing neonatal care, as timely interventions are vital for proper resuscitation.
However, current clinical methods for recording ToB often rely on manual processes, which can be prone to inaccuracies. 
In this study, we present a novel two-stream fusion system that combines the power of image and video analysis to accurately detect the ToB from thermal recordings in the delivery room and operating theater.
By integrating static and dynamic streams, our approach captures richer birth-related spatiotemporal features, leading to more robust and precise ToB estimation. 
We demonstrate that this synergy between data modalities enhances performance over single-stream approaches.
Our system achieves 95.7\% precision and 84.8\% recall in detecting birth within short video clips. 
Additionally, with the help of a score aggregation module, it successfully identifies ToB in 100\% of test cases, with a median absolute error of 2 seconds and an absolute mean deviation of 4.5 seconds compared to manual annotations.

\end{abstract}

\IEEEpeerreviewmaketitle

\section{Introduction}

\begin{figure*}[ht]
\centering
\begin{minipage}[b]{.97\linewidth}
  \centering
  \centerline{\includegraphics[width=\linewidth]{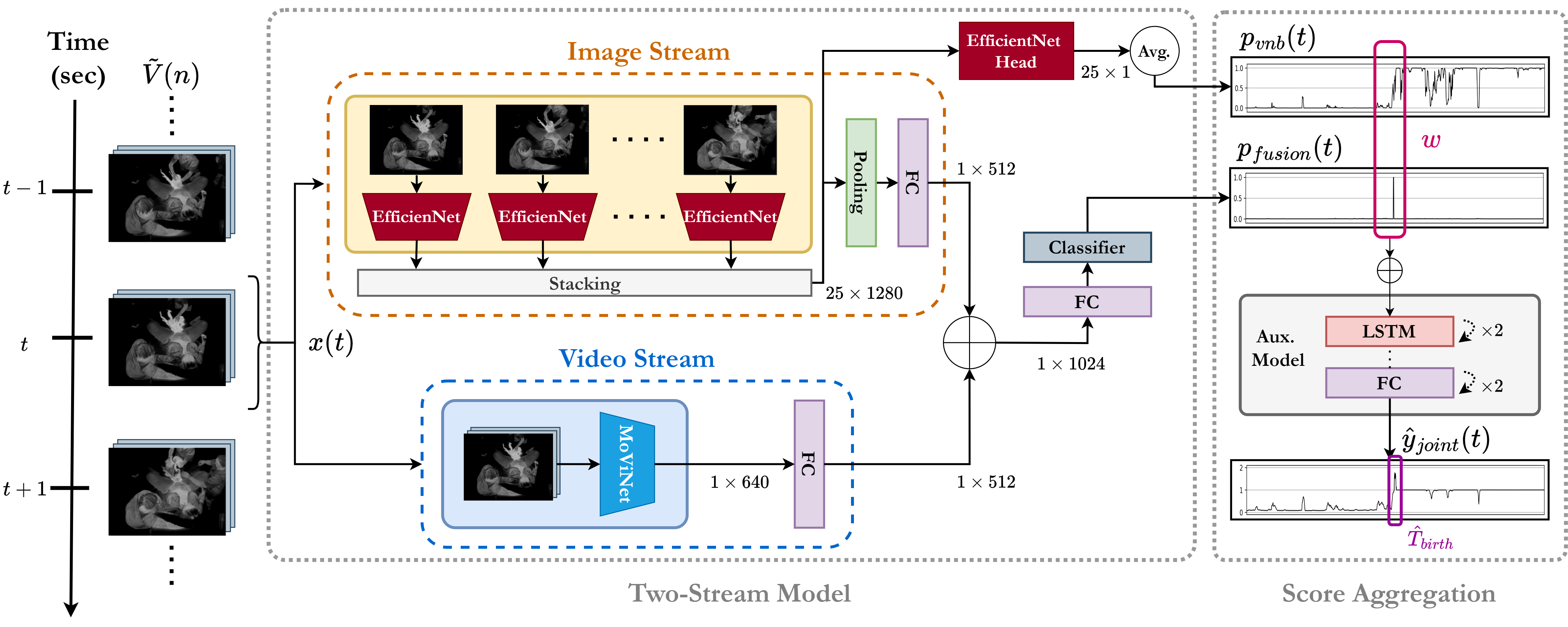}}
\end{minipage}
\caption{Overview of our proposed two-stream system for Time of Birth detection. At each second, a sliding window containing the previous 25 frames (3 seconds) is utilized as input $x(t)$ to our model. The video clip is processed independently in each stream by pretrained backbones, extracting static and temporal features that are concatenated to produce richer probability scores $p_{fusion}$. These scores pass through a score aggregation module where they are combined with image-specific scores $p_{vnb}$ and processed by an auxiliary model using a moving window $w$ to refine ToB predictions.}
\label{fig:overview}
\end{figure*}

Globally, 10\% of newborns require resuscitative interventions to start breathing at birth, with birth asphyxia causing nearly 900,000 deaths annually \cite{ersdal2012early, who-asphixia}. 
Neonatal resuscitation guidelines \cite{madar2021european, Greif2024} emphasize initiating interventions within the “golden minute” after birth, making accurate Time of Birth (ToB) documentation essential for quality improvement, documentation, debriefing, and potential real-time decision support \cite{branche2020first}.
However, ToB--when the newborn is fully visible outside the mother’s perineum--is typically recorded manually with minute precision, limiting the reliability of Newborn Resuscitation Algorithm Activities (NRAA) timelines for clinical use and evidence-based research. 
AI has shown promise in automating these timelines \cite{meinich2020activity, garcia-torres2023comparative}, but current approaches still depend on manually recorded ToB, restricting their potential.

The NewbornTime project \cite{engan2023newborn} seeks to bridge this gap by automating NewbornTimeline, a system designed to generate precise NRAA timelines, including ToB and resuscitation activities.
This enables large-scale analysis of newborn resuscitation videos, producing NRAA timelines that can refine medical guidelines and improve neonatal care.
A key innovation is the use of \emph{thermal imaging} for ToB detection, facilitating GDPR-compliant privacy protection while leveraging newborns' higher skin temperature for birth identification.
Earlier work in this project introduced the first AI-driven ToB detector \cite{Garcia-Torres2025}, which relied solely on static images. 
By continuously analyzing individual thermal frames, this method effectively detected the newborn's presence but sometimes lacked precision in identifying the exact moment of birth.
To improve accuracy, we later proposed a video-based system \cite{Garcia-Torres2025a} that extracted and classified short video clips from birth episodes.
This system captured dynamic movements and thermal changes throughout the video sequence, achieving more accurate ToB estimations.
However, this improvement came at the cost of an increased risk of missing detections.

In this study, building on our previous approaches \cite{Garcia-Torres2025, Garcia-Torres2025a}, we integrate static spatial analysis and motion pattern modeling into a unified system, combining the strengths of both methods.
Several studies \cite{Wang2018, Zolfaghari2018, Le2022, Tani2024, Yosry2024} have explored the fusion of image and video data for human activity recognition, demonstrating that it can enhance AI systems.
Here, we introduce a two-stream system for ToB detection that effectively processes both data modalities in a complementary manner.
By employing a fusion strategy that combines both feature-level and decision-level fusion, the system leverages information about the newborn’s visibility (image stream) and birth moment (video stream) in parallel to make more informed decisions. 
Our method improves accuracy in ToB detection while reducing the risk of missed detections, leading to more stable and reliable outcomes.
To the best of our knowledge, no other work has addressed automatic ToB detection.

\section{Data Material}
\label{sec:dataset}

The dataset comprises 611 thermal birth videos recorded semi-automatically \cite{Brunner2025} at Stavanger University Hospital (SUS), Norway. 
A passive thermal module by Mobotix \cite{mobotix:v1.02} was installed in eight delivery rooms and one operating theater to capture the recordings.
ToB was initially logged by midwives or nurse assistants using the Liveborn Observation App \cite{bucher2020digital}, a tool designed for documenting post-birth events in research settings, enabling more precise ToB recording than standard clinical practice \cite{kolstad2024detection}.  
After registering the ToB in Liveborn, a 30-minute thermal video (resolution: 252$\times$336, frame rate: 8.33 fps) is saved, covering 15 minutes before and after birth, while the remaining footage is discarded.

Due to the numerous tasks involved in childbirth, the registration of ToB in Liveborn was sometimes delayed.
To ensure accurate ToB records, manual annotation with second precision was done by reviewing the thermal videos.
258 videos also included an additional label indicating whether the newborn was clearly visible and identifiable in the video frames (Visible Newborn, VNB).
Video annotation was carried out by medical professionals using ELAN 5.8 (The Language Archive, Nijmegen, The Netherlands) \cite{wittenburg2006elan}.

\section{Methodology}
\label{sec:methodology}

We propose a unified system that leverages complementary visual cues from thermal imaging for ToB detection.
It involves three learning steps. 
We first train EfficientNet and MoViNet on birth episodes using their respective data modalities.
Next, we develop a two-stream model that integrates image and video patterns through feature-level fusion. 
Finally, we implement a score aggregation module that performs decision-level fusion to refine the final ToB detection.
This approach can be seen as a hybrid fusion where each step is learned independently.
An overview of our method is shown in Figure~\ref{fig:overview}.

\subsection{Preprocessing}
\label{subsec:vid_pp}

Previous work \cite{garcia2022towards} identified key challenges in using thermal sensors for ToB detection, highlighting the limitations of relying on absolute measurements. 
To address this, we proposed a Gaussian Mixture Model (GMM) normalization method \cite{Garcia-Torres2025} that models the temperature distribution around human skin temperature, enhancing consistency and preserving relative temperature variations.

Let $\mathcal{V} \in \mathbb{R}^{N \times H \times W}$ represent a single-channel thermal video consisting of $N$ frames of spatial resolution of $H \times W$. 
We define $I(n)$ as the thermal frame at index $n = 0,1,\dots,N-1$. A video clip $V(n)$ is then constructed using the preceding $F$ frames:
\begin{equation}
    V(n) = \{I(n),I(n-1),...,I(n-F+1)\}
\end{equation}
To normalize $\mathcal{V}$, a set of intensity values $\mathbf{v}$ are extracted and fitted to the GMM. 
The highest Gaussian mean $\hat{\mu}_\mathbf{v}$, selected by empirical constraints, is then used to define a temperature range of interest.
Our proposed normalization function $\phi_{\hat{\mu}_\mathbf{v}}$ clips the temperature values within this video-specific range and rescales them to $[0,1]$, resulting in a normalized video $\Tilde{\mathcal{V}} = \phi_{\hat{\mu}_\mathbf{v}}(\mathcal{V})$.

To construct video clips, frames are sampled at defined time intervals, where the frame index $n$ is mapped to a timestamp $t$ using the frame rate $f_r$ and a temporal stride $\tau$. 
The final sampled clip set $x(t)\in\mathbb{R}^{F\times H\times W}$ is given by:
\begin{equation} 
    x(t) = \Tilde{V}(\lfloor f_r \cdot t \rfloor), \quad t=t_0,t_0+\tau,t_0+2\tau,...,T
\end{equation}
where $\lfloor \cdot \rfloor$ is the floor function, $t_0 = \lfloor F / f_r \rfloor$ is the first valid timestamp, and $T = \lfloor N/f_r \rfloor$ is the total video duration.
Values of $t$ before $t_0$ are excluded to avoid boundary problems.


\subsection{Two-Stream Architecture}
\label{subsec:two-stream}

To effectively capture both spatial and temporal information, we propose a two-stream neural network that integrates static image-based analysis and video-based temporal modeling through a feature-level fusion strategy. 
At every time step $t$, the model receives a video clip $x(t)$ containing the preceding $F=25$ frames (3 seconds). 
The image stream processes individual thermal frames to detect the presence of the newborn within the clip.
Frame-level features are then stacked across the clip and averaged to form a global clip representation.
In parallel, the video stream evaluates the entire clip, providing contextual information about the birth beyond isolated images.

We employ EfficientNet \cite{Tan2021} and MoViNet \cite{kondratyuk2021movinets} as feature extractors (marked in red and blue in Figure~\ref{fig:overview}) due to their strong capabilities and efficiency. 
Each stream is followed by a fully connected (FC) layer to refine the learned embeddings.
The outputs from both streams are concatenated into a shared representation, passed through an additional FC layer, and then processed by a binary classifier, implemented as another FC layer,
resulting in a fusion score $p_{fusion}$.
Additionally, a VNB score $p_{vnb}$ is independently obtained by processing the stacked image-specific features through the EfficientNet head and averaging the resulting probability scores.

When training the two-stream model, we adopt a transfer learning approach, freezing both the backbones and the EfficientNet head. 
Only the added top layers are trained.

\subsection{Score Aggregation}
\label{subsec:inference}

Instead of relying on finite impulse response (FIR) filtering and a high-confidence threshold (HCT) postprocessing, as proposed in \cite{Garcia-Torres2025} and \cite{Garcia-Torres2025a}, we introduce a score aggregation module designed to capture richer patterns and improve ToB detection accuracy.
The purpose of this module is to use the image stream as both a fallback mechanism and a reference to refine estimations.
After processing an entire video through our two-stream model, this module combines the resulting $p_{fusion}(t)$ with the VNB scores $p_{vnb}(t)$.
Both scores are stacked into a $T \times 2$ matrix and processed in overlapping segments using a 10-second moving window $w$ with a stride of 1 second.
Each segment is then passed through an auxiliary lightweight neural network comprising two Long Short-Term Memory (LSTM) \cite{Hochreiter1997} and two FC layers.

To guide the learning process of the auxiliary model, we define a custom loss function $\mathcal{L}$ incorporating three supervision signals used as labels.
The event signal $y_{evt}(t) \in \{0,1\}$ is a sparse signal with zeros except for a peak at ToB, indicating the birth moment.
The transition signal $y_{tr}(t) \in \{0,1\}$ marks the transition from pre-birth frames (zero) to post-birth frames (one).
The joint signal $y_{joint}(t) \in \{0,1,2\}$ provides a unified and more comprehensive supervision signal, computed as the sum of $y_{evt}(t)$ and $y_{tr}(t)$.
The auxiliary model estimates $\hat{y}_{evt}(t)$ and $\hat{y}_{tr}(t)$ via independent sigmoid functions, while $\hat{y}_{joint}(t)$ is computed as their linear combination.
Binary cross-entropy (BCE) \cite{bishop2006pattern} is used for the event and transition labels, while mean squared error (MSE) is applied to the joint label.
The total loss is formulated as:
\begin{equation}
\label{eq:loss}
\begin{aligned}
    \mathcal{L} = & \ \alpha_1\text{BCE}(y_{evt}, \hat{y}_{evt}) + \alpha_2\text{BCE}(y_{tr}, \hat{y}_{tr}) \\
    & \ + \alpha_3\text{MSE}(y_{joint}, \hat{y}_{joint})
\end{aligned}
\end{equation}
where $\alpha_1$, $\alpha_2$, and $\alpha_3$ are weighting coefficients.  
By integrating these labels, the auxiliary model learns not only to pinpoint the ToB but also to recognize the transition dynamics before and after birth.
The joint estimation $\hat{y}_{joint}(t)$ serves as a linear projection of the learned patterns, leading to more robust predictions compared to using FIR and HCT. 
Finally, the estimated ToB $\hat{T}_{birth}$ is given by:
\begin{equation}
    \hat{T}_{birth} = \arg\max_{t} \hat{y}_{joint}(t) \quad \text{if } \max\hat{y}_{joint}(t)\geq\gamma
\end{equation}
The decision threshold $\gamma$ is set to 0.95, representing the midpoint of the expected range $[0,2]$ and accounting for variations within a 5\% uncertainty.

\section{Experiments}
\begin{figure*}[t]
\centering
\begin{minipage}[b]{\linewidth}
  \centering
  \centerline{\includegraphics[width=\linewidth]{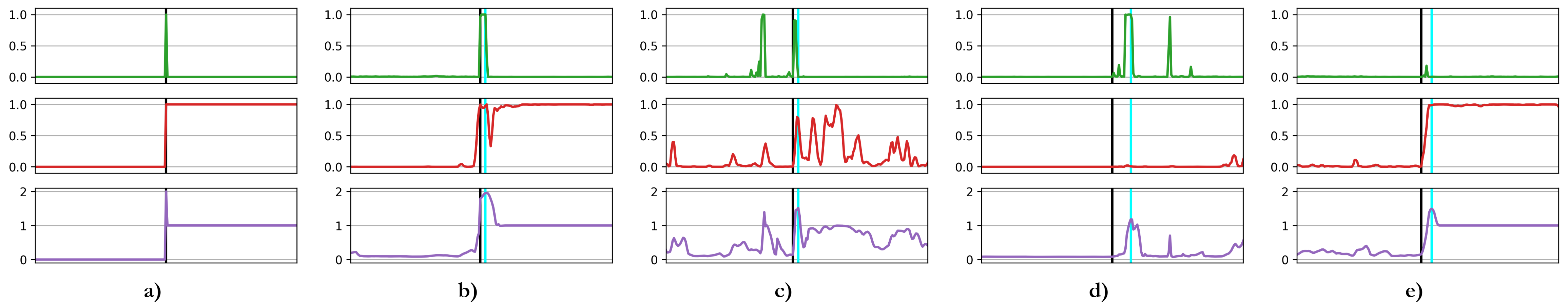}}
\end{minipage}
\caption{Examples of predicted event (green), transition (red), and joint (purple) signals around ToB. $T_{birth}$ is depicted in black, and $\hat{T}_{birth}$ in cyan. $a)$ Ground truth signals. $b)$ Event and transition signals correctly identify the ToB. $c)$ The auxiliary model avoids false positives by finding the best trade-off between event and transition signals. $d) \text{ and } e)$ If any signal fails, the auxiliary model relies on the other signal to estimate ToB, acting as a fallback system.}
\label{fig:results}
\end{figure*}

The NewbornTime project is still collecting data, and the dataset has expanded since our previous studies, now including 258 videos for the image-based task with VNB annotations and 611 for the video-based task.
As a result, we first retrain our ToB detectors with this updated dataset, maintaining the same methodology with minor hyperparameter refinements.
We also include operation theater videos for training MoViNet, which was previously trained solely on delivery room videos in \cite{Garcia-Torres2025a}.
For testing, 35 thermal videos (25 from delivery rooms and 10 from operating theaters) are selected.
All videos contain ToB annotation, with 19 also including extra labels for the image-based task: VNB and No Newborn (NNB). 
The remaining videos are split into training and validation sets (85\%/15\%).

We train the two-stream model following the approach used in \cite{Garcia-Torres2025a}, inputting short video clips into the model. 
Training is performed for up to 100 epochs with early stopping. 
The Adam optimizer is used with $\beta_1$=0.9, $\beta_2$=0.999, and a learning rate of 0.0001, decaying exponentially by a factor of 0.97 each epoch.
Two Tesla V100 GPUs with 32 GB RAM are utilized with a batch size of 8 per GPU.
Data augmentation includes brightness/contrast adjustments and random left-right flipping for enhanced generalization.

The auxiliary model in the score aggregation module was trained using the validation set employed when training the two-stream model.
The model is trained for 100 epochs with early stopping, using Adam with an initial learning rate of 0.001 and exponential decay of 0.97 per epoch.
The weights in Eq.~\ref{eq:loss} are empirically set to $\alpha_1 = \alpha_2 = 0.4$ and $\alpha_3 = 0.2$, prioritizing the learning of event and transition functions.

Evaluation is done using precision and recall for binary classification \cite{lipton2014optimal}, and Matthew's Correlation Coefficient (MCC) \cite{chicco2020advantages} as a more comprehensive metric.
Additionally, we define the error $err$ as the time difference between $\hat{T}_{birth}$ and the manually annotated ToB ($T_{birth}$):
\begin{equation}
    err = \hat{T}_{birth} - T_{birth}
\end{equation}
A positive error indicates a late prediction, while a negative one indicates an early prediction.
We also compute statistical measures (quartiles and mean) using the absolute error $|err|$.

\section{Results \& Discussion}

Table~\ref{tab:model_eval} and Table~\ref{tab:tob_est} present the classification performance and ToB estimation evaluation, respectively.
Since image and video modalities differ in task complexity, input data, and evaluation criteria, their classification results in Table~\ref{tab:model_eval} are not directly comparable; the focus should be on improvements within each modality.

Retraining our previous ToB detectors led to independent performance improvements in both models within their respective data domains.
EfficientNet clearly benefits from the increased data, particularly in reducing false positives.
When evaluating the 35 test videos with the MoViNet model from \cite{Garcia-Torres2025a}, performance dropped due to its lack of training on operation theater videos, though precision remained high.
The retrained MoViNet preserves the same level of precision while significantly improving recall, leading to more birth estimations but with increased variability (reflected in a higher mean value in Table~\ref{tab:tob_est}).
This behavior may result from the inherent data imbalance in our dataset.
Besides, training becomes even more challenging due to the added complexity of handling both delivery room and operation theater environments.

\begin{table}[t]
\centering
\caption{Performance on classification. VNB and ToB serve as the positive classes for image and video classification. The number of test videos varies based on available annotations.}
\begin{tabular}{cclcccc}
\hline
Modality & Test Size & Model & Precision & Recall & MCC \\ \hline \hline
\multirow{2}{*}{Image} & \multirow{2}{*}{19} & EfficientNet\cite{Garcia-Torres2025} & 0.867 & 0.946 & 0.881 \\
& & EfficientNet - retrained & 0.945 & 0.938 & 0.927 \\ \hline \hline
\multirow{3}{*}{Video} & \multirow{3}{*}{35} & MoViNet \cite{Garcia-Torres2025a} & 0.944 & 0.648 & 0.774 \\
& & MoViNet - retrained & 0.934 & 0.81 & 0.864 \\
& & \textbf{Two-Stream} & \textbf{0.957} & \textbf{0.848} & \textbf{0.896} \\ \hline \hline
\end{tabular}
\label{tab:model_eval}
\end{table}

Our two-stream model benefits from the combination of static and dynamic features, achieving better video clip classification performance than MoViNet.
However, this result does not directly translate to better ToB estimations in Table~\ref{tab:tob_est}.
In \cite{Garcia-Torres2025} and \cite{Garcia-Torres2025a}, the strict postprocessing method--combining FIR and HCT--proved highly effective in detecting the ToB by largely reducing false positives during inference.
Interestingly, our two-stream model achieves strong results even without postprocessing, relying solely on the fusion scores $p_{fusion}$ with binary thresholding at 0.5 and first-occurrence selection.
This approach provides an absolute median error of just 1 second but exhibits high variability, as reflected in the Q3 and mean values.
Applying FIR and HCT to $p_{fusion}$ slightly improves accuracy but leads to more missed birth detections.

Finally, we find that the score aggregation module acts both as a fallback system and guidance mechanism, helping to find meaningful patterns between $p_{fusion}$ and $p_{vnb}$ to refine ToB estimation, as depicted in Figure~\ref{fig:results}.
By combining both probability scores and leveraging the auxiliary model, birth detection improves significantly, achieving an absolute median error of 2 seconds with low variability while detecting 100\% of birth events.
Moreover, integrating the two-stream network with the score aggregation module adds minimal computational overhead compared to running MoViNet and EfficientNet independently (see Table~\ref{tab:eff_analysis}), making the system suitable for real-time use.
While the proposed method is retrospective, in real-time, the aggregation module would process a fixed-length window of available data.

We explored various feature fusion strategies in our two-stream model, including weighted attention and self-attention, but all yielded similar performance.
We suspect that the limiting factor was the quality of the video and image streams, as misclassifications consistently occurred in the same thermal videos--specifically, those where birth visibility or newborn appearance was reduced.
Thus, we opted for the simplest approach.
We also tested different architectures for the score aggregation module, reporting only the best-performing configuration.
Using VNB annotations as ground truth instead of the transition signal could better support the auxiliary model, but the lack of annotations makes this infeasible for now.

\begin{table}[t]
\centering
\caption{Statistics on ToB detection for the 35 test videos (values in seconds). Q-values represent quartiles. Count denotes the percentage of test videos where $\hat{T}_{birth}$ was found.}
\begin{tabular}{lccccc}
\hline
\multicolumn{1}{c}{Model} & Q1 & Q2 & Q3 & Mean & Count\\ \hline \hline
EfficientNet \cite{Garcia-Torres2025} (FIR + HCT)  & 1.6 & 2.7 & 26.6 & 71.8 & 100\% \\
EfficientNet - retrained (FIR + HCT)  & 1.2 & 3.1 & 8.1 & 53.8 & 100\% \\ \hline \hline
MoViNet \cite{Garcia-Torres2025a} (FIR + HCT) & 1 & 2 & 3 & 2.2 & 71.4\% \\
MoViNet - retrained (FIR + HCT) & 1 & 1 & 4 & 17.1 & 82.9\% \\ \hline \hline
Two-Stream  & 0 & 1 & 19 & 63.5 & 94.3\% \\ 
Two-Stream (FIR)  & 0 & 1 & 5 & 17 & 85.7\% \\ 
Two-Stream (FIR + HCT)  & 1 & 1 & 2.5 & 14.9 & 77.1\% \\ 
\textbf{Two-Stream + Agg.}  & \textbf{1} & \textbf{2} & \textbf{3} & \textbf{4.5} & \textbf{100\%} \\ \hline \hline
\end{tabular}
\label{tab:tob_est}
\end{table}

\begin{table}[t]
\centering
\caption{Computational efficiency analysis. Parameter count (in millions) includes both the backbone and classifier. The average time (in milliseconds) per single inference step using one V100 GPU is measured.}
\begin{tabular}{lcccccc}
\hline
\multicolumn{1}{c}{Model} & Params (M) & Method & Running Time (ms) \\ \hline \hline
EfficientNet & 8.2 & Frame-wise & 23 \\
MoViNet & 6.1 & Clip-wise & 75 \\
Two-Stream & 14.6 & Clip-wise & 104 \\ \hline \hline
\end{tabular}
\label{tab:eff_analysis}
\end{table}

\section{Conclusion}

In this work, we introduce an approach for automatic ToB detection by fusing spatial analysis and temporal modeling into a unified two-stream system, capturing both static and dynamic visual cues from thermal imaging. 
A key innovation is the score aggregation module, which refines ToB estimation by enabling the auxiliary model to learn complex spatial-temporal patterns. 
Together, these components enhance accuracy and reduce missed detections, ensuring more reliable and robust outcomes.
Our system detects ToB in 100\% of test cases, with an absolute median deviation of 2 seconds and an absolute mean deviation of 4.5 seconds from manual annotations.

Future work will focus on refining the models with the growing dataset and exploring alternative fusion strategies to enhance ToB detection. 
With a larger, fully annotated dataset, we aim to investigate an end-to-end approach to jointly train the two-stream model and aggregation module, optimizing their synergy within the system.
Additionally, we will finalize NewbornTimeline, an automated system for generating detailed NRAA timelines, including ToB detection and newborn resuscitation activity recognition.


\section*{Acknowledgment}

The NewbornTime project is funded by the Norwegian Research Council (NRC), project number 320968. Additional funding has been provided by Helse Vest, Fondation Idella, and Helse Campus, Universitetet i Stavanger. Study registered in ISRCTN Registry, number ISRCTN12236970.

\bibliographystyle{IEEEtran}
\bibliography{strings}

\end{document}